\documentclass[runningheads]{llncs}
\usepackage[T1]{fontenc}
\usepackage{graphicx}
\usepackage{booktabs}
\usepackage[misc]{ifsym}
\newcommand{\corr}{(\Letter)}

\usepackage{eurosym}
\usepackage{makecell}
\usepackage{amssymb}
\usepackage[hyphens]{url}
\usepackage{hyperref}
\usepackage{subcaption}
\usepackage{longtable}
\usepackage{multirow}

\begin{document}

\title{On the Performance of LLMs\\ for Real Estate Appraisal}

\author{Margot Geerts \inst{1} \corr
\and Manon Reusens\inst{1}
\and Bart Baesens\inst{1,3}
\and Seppe {vanden Broucke}\inst{2,1}
\and Jochen {De Weerdt}\inst{1}}

\authorrunning{M. Geerts et al.}

\institute{LIRIS, KU Leuven, Leuven, Belgium \email{margot.geerts@kuleuven.be}
 \and
 Department of Business Informatics and Operations Management, Ghent University, Ghent, Belgium
\and
Department of Decision Analytics and Risk, University of Southampton, Southampton, UK}

\tocauthor{Margot Geerts, Manon Reusens, Bart Baesens, Seppe vanden Broucke, Jochen De Weerdt}
\toctitle{On the Performance of LLMs for Real Estate Appraisal}

\maketitle              %

\begin{abstract}
The real estate market is vital to global economies but suffers from significant information asymmetry. This study examines how Large Language Models (LLMs) can democratize access to real estate insights by generating competitive and interpretable house price estimates through optimized In-Context Learning (ICL) strategies. We systematically evaluate leading LLMs on diverse international housing datasets, comparing zero-shot, few-shot, market report-enhanced, and hybrid pro-mpting techniques. %
Our results show that LLMs effectively leverage hedonic variables, such as property size and amenities, to produce meaningful estimates. While traditional machine learning models remain strong for pure predictive accuracy, LLMs offer a more accessible, interactive and interpretable alternative. Although self-explanations require cautious interpretation, we find that LLMs explain their predictions in agreement with state-of-the-art models, confirming their trustworthiness. Carefully selected in-context examples based on feature similarity and geographic proximity, significantly enhance LLM performance, yet LLMs struggle with overconfidence in price intervals and limited spatial reasoning.
We offer practical guidance for structured prediction tasks through prompt optimization. Our findings highlight LLMs’ potential to improve transparency in real estate appraisal and provide actionable insights for stakeholders.

\end{abstract}

\section{Introduction}
Global real estate, valued at \$379.7 trillion in 2022, represents the world's largest wealth store, with residential properties constituting the majority\footnote{ \url{https://www.savills.com/impacts/market-trends/the-total-value-of-global-real-estate-property-remains-the-worlds-biggest-store-of-wealth.html}}. The real estate market plays a crucial role in economies worldwide, impacting homeowners, investors, and governments. Accurate price estimations are vital for all stakeholders, from home buyers facing affordability challenges in Europe\footnote{ \url{https://ec.europa.eu/eurostat/web/interactive-publications/housing-2023}} and the U.S.\footnote{ \url{https://www.forbes.com/advisor/mortgages/real-estate/housing-market-predictions/}} to China's slowing market\footnote{ \url{https://www.imf.org/en/News/Articles/2024/02/02/cf-chinas-real-estate-sector-managing-the-medium-term-slowdown}}. Access to reliable price data helps ensure informed decision-making and supports a stable and sustainable market across regions. Nevertheless, real estate valuation remains opaque and unevenly accessible, contributing to information asymmetry between buyers and sellers~\cite{kurlat2015}. Sellers inherently have superior knowledge of the local market and the property's condition as opposed to buyers. While potential buyers can call upon a real estate broker or other experts, this asymmetry is difficult to eliminate. Some argue that a data-driven house price prediction approach can help real estate stakeholders, including buyers, by informing their decisions \cite{Bastos2024,Li2024}. However, this approach requires advanced Machine Learning (ML) expertise, extensive manual data processing and access to a substantial dataset, which may not be readily available to the average home buyer. Large Language Models (LLMs) present a promising solution to address this information asymmetry \cite{weiss2024redesigning}. Trained on vast and diverse datasets encompassing a significant portion of Internet knowledge \cite{dubey2024llama3herdmodels}, these models have the potential to uncover meaningful insights and patterns, including those relevant to real estate \cite{Gloria2024}. Recently, LLMs have been proven to excel in structured prediction tasks with In-Context Learning (ICL), enabling them to approximate regression problems without explicit training \cite{vacareanu2024words}. This makes them a promising tool for ad hoc house price prediction, reducing the barriers to data-driven real estate insights. By leveraging their extensive training, LLMs could bridge knowledge gaps, offering nuanced perspectives and data-driven guidance in this complex domain. This marks a key step towards democratizing access to real estate appraisal insights and enhancing transparency for a diverse range of stakeholders. Reducing information asymmetry improves price accuracy, benefiting both buyers, who avoid overpaying, and sellers, who experience faster sales due to improved liquidity and as such receiving fair market value \cite{kurlat2015}. Additionally, investors, financial institutions, and policymakers can make more informed decisions, leading to improved investment strategies, risk assessments, and more effective tax and policy frameworks.

In this paper, we assess LLMs' potential for improving accessibility to real estate appraisal, or valuation, by answering four Research Questions (RQs): \begin{enumerate}
    \item [\textbf{RQ1}] How effectively can prompt engineering techniques optimize LLM performance for house price prediction and what is the most effective prompt?
    \item [\textbf{RQ2}] Can LLMs generate sufficiently accurate house price estimates to serve as viable alternatives to traditional ML models?
    \item [\textbf{RQ3}] How reliably do LLMs estimate price intervals for real estate appraisal, and how does this compare to traditional ML approaches?
    \item [\textbf{RQ4}] What features do LLMs prioritize in their house price prediction processes, and how do these align with traditional valuation methodologies?
\end{enumerate}

To address these questions, we investigate different prompting approaches with ICL and evaluate a wide range of pre-trained LLMs on various housing datasets worldwide. In the context of the house price prediction task, we scrutinize the capabilities of LLMs in three dimensions: the accuracy of price predictions, the delineation of price intervals, and their explanatory capacity. Our contributions can be summarized as follows:

\begin{enumerate}
    \item We demonstrate that optimizing prompt design significantly improves LLM performance in house price prediction. Carefully selecting in-context examples based on feature similarity and geographic proximity enhances accuracy and adaptability across different housing markets.
    \item We show that LLMs can generate sufficiently accurate house price estimates, approaching the performance of traditional ML models. While they do not surpass ML models in predictive accuracy, their accessibility, interpretability, and flexibility make them valuable for real estate stakeholders.
    \item We identify overconfidence in price intervals as a key limitation of LLM-based valuation. LLMs consistently underestimate price uncertainty, producing narrower prediction ranges that fail to capture real market values.
    \item We find that LLMs prioritize hedonic property features effectively but struggle with spatial and temporal reasoning. Despite leveraging variables like property size and amenities, they undervalue the role of location and time.
    \item We help reduce information asymmetry in the real estate market by providing concrete guidelines for harnessing LLMs to support informed decision-making among buyers, sellers, financial institutions and policymakers.
\end{enumerate}

\section{Related Work}
\paragraph{House price prediction} is typically framed as a supervised learning problem involving tabular data. Automated Valuation Models (AVMs) are trained on datasets $D=\{(X_i, y_i)\}^n_{i=1}$ where $X_i \in {\rm I\!R}^m$ are the $m$-dimensional features of property $i$, and $y_i \in {\rm I\!R}$ is its price. The objective is to minimize prediction error $L(\hat{y},y)$, using loss functions like mean squared error (MSE). Features are broadly categorized into hedonic attributes, i.e. structural attributes (e.g., size, number of rooms), and locational factors (e.g., coordinates, proximity to amenities). Recent research has shifted from hedonic regression models \cite{Bourassa2003} to modern ML and deep learning approaches \cite{Geerts2023,Lee2023,Li2024}, with increasing focus on interpretability, including Shapley values \cite{Rico-juan2021} and uncertainty quantification techniques such as conformal prediction for prediction intervals \cite{Bastos2024,pmlr-v230-hjort24a}.

\paragraph{LLMs for data science} have revolutionized predictive modeling with unstructured textual data. In house price prediction, Natural Language Processing (NLP) extracts insights from property descriptions, market reports, and reviews, converting them into structured features \cite{Shen2020,Zhang2024}. Building on recent advancements in NLP, pre-trained LLMs are increasingly used in data science applications with tabular data \cite{wu-hou-2025-efficient}, such as geospatial interpolation \cite{manvi2024geollm}, Point of Interest recommendation \cite{Feng2024}, and time series analysis \cite{jin2024llmbasedknowledgepruningtime}. Despite their growing adoption, research on LLMs for real estate appraisal remains limited, with a recent study focusing only on rental price prediction \cite{Chen2024}. Our work addresses this gap by examining more robust prompting strategies, evaluating diverse datasets, and integrating interpretability, thereby offering new insights into real estate appraisal with LLMs.

\paragraph{In-Context Learning (ICL)} is an inference-time technique where the model, without updating its parameters, generalizes from provided examples. Specifically, LLMs are provided with K examples as context and are then tasked with completing a new example by leveraging the patterns and information from the preceding ones \cite{manikandan2023language}. \cite{vacareanu2024words} show how LLMs can perform regression tasks when provided in-context examples. LLMs' ability to handle tabular data is demonstrated through frameworks like TabLLM for data-efficient classification \cite{pmlr-v206-hegselmann23a}, while the Meta-ICL framework further enhances ICL efficiency \cite{Coda-Forno2023}.

\section{Methodology}

\subsection{Large Language Models for House Price Prediction}

To effectively prompt LLMs with tabular housing data, we follow the guidelines set by \cite{pmlr-v206-hegselmann23a} for manual data serialization in zero- and few-shot learning settings. Our goal is not to replace traditional ML models but to evaluate LLMs' capabilities in estimating house prices and identify the optimal prompt.
\begin{figure}[ht]
    \centering
    \includegraphics[width=1\linewidth]{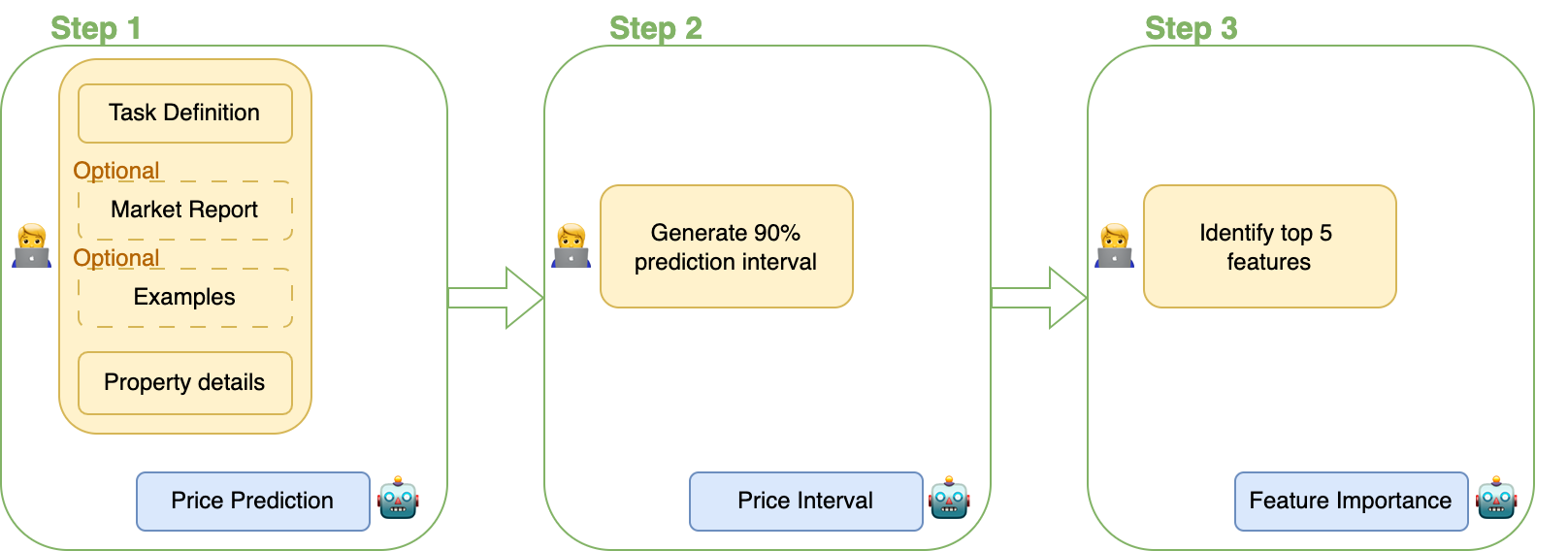}
    \caption{Overview of the LLM prompting methodology for house price prediction. Step 1: The model receives a structured prompt containing the task definition, an optional market report, optional ICL examples, and details of the target property-forming the basis for prompt optimization. It then predicts the property price. Step 2: The model generates a 90\% prediction interval. Step 3: The model identifies the five most important features. This approach enables price estimation, uncertainty quantification, and interpretability in real estate appraisal.}
    \label{fig:methodology}
\end{figure}

Figure \ref{fig:methodology} illustrates the prompting strategy used in the experiments. To determine the optimal prompt for house price prediction, we evaluate twelve different strategies combining various building blocks. Specifically, we incorporate market reports and in-context examples on top of the zero-shot baseline containing the task definition and property description. Our prompting optimization strategy uses ICL to enforce two pillars that are relevant in real estate: regional real estate market dynamics, which are often accounted for in AVMs by explicitly modeling temporal effects \cite{Lee2023}, and comparable property valuation examples \cite{Li2024}, essentially delineating housing submarkets \cite{Bourassa2003}. The market reports provide context on regional house price indices from the preceding month or quarter. The labeled examples from the training data are selected based on either haversine distance (geographic proximity) or cosine distance (similar hedonic features) and limited to three or ten examples. Finally, we test a combination of ten examples evenly split between geographic and hedonic neighbors. This combination of geographic and characteristic-based similarity has been shown to be effective in hedonic price modeling in prior work \cite{Ozhegov2022}. Based on these results, we extend the interaction with the LLM by maintaining the conversation history. Using the best-ranked prompt configuration across datasets and LLMs, the LLM generates a price interval with a 90\% target coverage. While methods like conformal prediction require access to the internal LLM structure, which is unavailable in standard API interactions, direct prompting aligns with how real estate practitioners would use LLMs. Finally, we assess explainability by asking the model to identify the top five features influencing its prediction. The full prompt template is shown in Appendix \ref{a1}.

 Since LLMs struggle with raw geographic coordinates due to tokenization of prompts, we follow \cite{manvi2024geollm} in reverse geocoding the coordinates into full addresses using the Nomatim API\footnote{ \url{https://nominatim.org/}} to OpenStreetMap\footnote{\url{https://www.openstreetmap.org/}}, incorporating both representations into the prompt.

Given the importance of accessibility in this study, we include a range of the most recent pre-trained LLMs comprising both open-source and closed-source models of varying sizes and architectural design. In our experiments, we use Llama 3.2:3B \cite{Meta2024}, Llama 3.1:70B \cite{dubey2024llama3herdmodels}, and GPT-4o-mini \cite{OpenAI2024}. These models were selected to balance scale (number of parameters), provider diversity, and accessibility constraints. All models were prompted with a seed of 0 and temperature of 0 to ensure reproducibility. All code and data used for the experiments is available via \url{https://github.com/margotgeerts/LLM4RealEstate}. More information on the checkpoints and computing environment used can be found in Appendix \ref{a2}.

\subsection{ML baselines} 
We consider two common house price prediction baselines: k-Nearest Neighbor (kNN) regression and Gradient Boosted Trees (GBT). These baselines are chosen based on their conceptual relevance to our LLM-approach and their strong empirical performance in real estate appraisal.
First, kNN serves as a natural baseline because it predicts a property's price based on an interpolation of its nearest neighbors. This aligns well with our LLM prompting strategy, which provides the model with similar properties as context. Comparing LLMs to kNN allows us to determine whether LLMs can extract deeper insights beyond simple interpolation. To ensure a fair comparison, we match the LLM prompt settings with $k=\{3,10\}$ neighbors, using haversine distance (geographic proximity), cosine distance (hedonic similarity), or a combination of both for ten examples. Second, we include GBTs—specifically LightGBM (LGBM) \cite{Ke2017}—as they remain the state-of-the-art (SOTA) choice for structured tabular data and have consistently outperformed deep learning methods in house price prediction tasks \cite{Geerts2024}. Unlike kNN, LGBM learns complex, nonlinear relationships in data, allowing us to benchmark whether LLMs can approach fully optimized ML models that have access to structured training data. We use LGBM with default parameters to ensure a fair, out-of-the-box comparison that mirrors how practitioners might deploy an ML model without extensive hyperparameter tuning. Other ML methods, such as linear regression or support vector machines (SVMs), were excluded as they generally underperform compared to GBTs on tabular data. Similarly, deep learning models such as Graph Neural Networks, while promising, have not yet demonstrated consistent superiority over GBTs in house price prediction \cite{Geerts2024}.

To compare prediction intervals between LLMs and LGBM, we use Conformal Prediction (CP). CP is a framework that provides valid prediction intervals without assuming any specific model, offering a distribution-free method for uncertainty quantification in ML tasks \cite{Vovk2005}. This approach works by using the observed data to ``conform'' the model’s predictions, ensuring that the true value lies within the predicted interval with a specified confidence level. Since house prices exhibit temporal trends that violate the assumption of data exchangeability (i.e., the data distribution is not independent and identically distributed over time), we apply a CP procedure designed to handle such distribution shifts, known as EnbPI \cite{xu2023conformalpredictiontimeseries}, via the MAPIE library \cite{taquet2022mapieopensourcelibrarydistributionfree}. 
Finally, we use the SHapley Additive exPlanations (SHAP) values \cite{Lundberg2017} to compare the LLM and LGBM explanations. Considering that the LLMs are provided with both coordinates and address, we adjust for this by aggregating the SHAP values corresponding to the two coordinate features (X-Y). Subsequently, we rank all features based on the mean absolute SHAP value computed across the test instances.

\subsection{Evaluation metrics}
To evaluate predictive performance, we report the Mean Absolute Percentage Error (MAPE), consistent with prior work \cite{Li2024}, due to space limitations, and the standard deviation of the Percentage Error (PE) across test observations. 
Prediction intervals are assessed based on actual coverage (percentage of instances where the true price is within the predicted interval) and the Mean Prediction Interval Width (MPIW), which measures precision. Valid intervals should achieve coverage close to 90\%, with narrower MPIWs indicating higher precision. Feature importance is compared by evaluating the top five features prioritized by LLMs and SHAP-based rankings from LGBM. With this comparison, we do not attempt to evaluate the ground truth correctness of these explanations. Our objective is to examine the degree of alignment between two distinct paradigms: LLMs and GBTs. Validating the intrinsic accuracy or faithfulness of the LLM explanations would require expert assessments or adversarial testing, which guide important directions for future work.

\subsection{Datasets}
To generalize LLM performance for real estate appraisal, we selected four real-world housing datasets located in various geographic areas. The datasets from King County, USA (from Kaggle\footnote{  \url{https://www.kaggle.com/datasets/astronautelvis/kc-house-data}}), Flanders, Belgium (proprietary), and Beijing, China (from Kaggle\footnote{ \url{https://www.kaggle.com/datasets/ruiqurm/lianjia}}) contain property transactions, while the dataset from Barcelona, Spain \cite{rey-blanco2024} contains property listings. Despite the subtle distinction between listings and transactions, we treat them similarly. A 60:20:20 train-validation-test split is used, and results are reported based on a random subset of 1000 test examples per dataset. Table \ref{tab:description} summarizes key statistics.

\begin{table}[ht]
\caption{Summary statistics of the housing datasets used for evaluation.}
    \label{tab:description}
    \centering
    \begin{tabular}{l|c|c|c|c}
    \toprule
         & King County & Flanders & Barcelona & Beijing \\ \midrule
        Train size & 12914 & 174135&25714 & 117349 \\
        Validation size & 4349& 59188& 12339& 37045\\
        Test size & 3569&55125 &23295 &41301\\
        No. variables &14 & 16&35& 17\\
        Min. price &\$75 000 &\euro34 280& \euro37 000 & \textyen1 270 021 \\
        Max. price & \$7 700 000&\euro970 512& \euro4 866 000& \textyen11 000 094\\
        Min. date & 2014-05-02& 2015-01-04&2018-03-01 & 2015-01-01\\
        Max. date & 2015-05-27& 2023-05-24& 2018-12-01& 2017-12-31\\
        \bottomrule
    \end{tabular}
\end{table}

\section{Results \& Discussion}
This section examines the effectiveness of large language models (LLMs) in house price prediction, focusing on both prompt optimization and comparisons with traditional ML approaches. First, we analyze the impact of different prompt engineering strategies on LLM performance, identifying the most effective techniques for improving prediction accuracy (\textbf{RQ1}). Next, we compare LLM-based predictions with traditional ML baselines, evaluating their absolute accuracy, interval estimates, and feature prioritization (\textbf{RQ2}–\textbf{RQ4}). Finally, we synthesize these findings into practical guidelines, offering insights on when and how LLMs can be effectively deployed for real estate appraisal.

\subsection{Optimizing LLMs with Prompt Engineering}
\begin{figure}[h]

    \centering
    \includegraphics[width=0.9\linewidth]{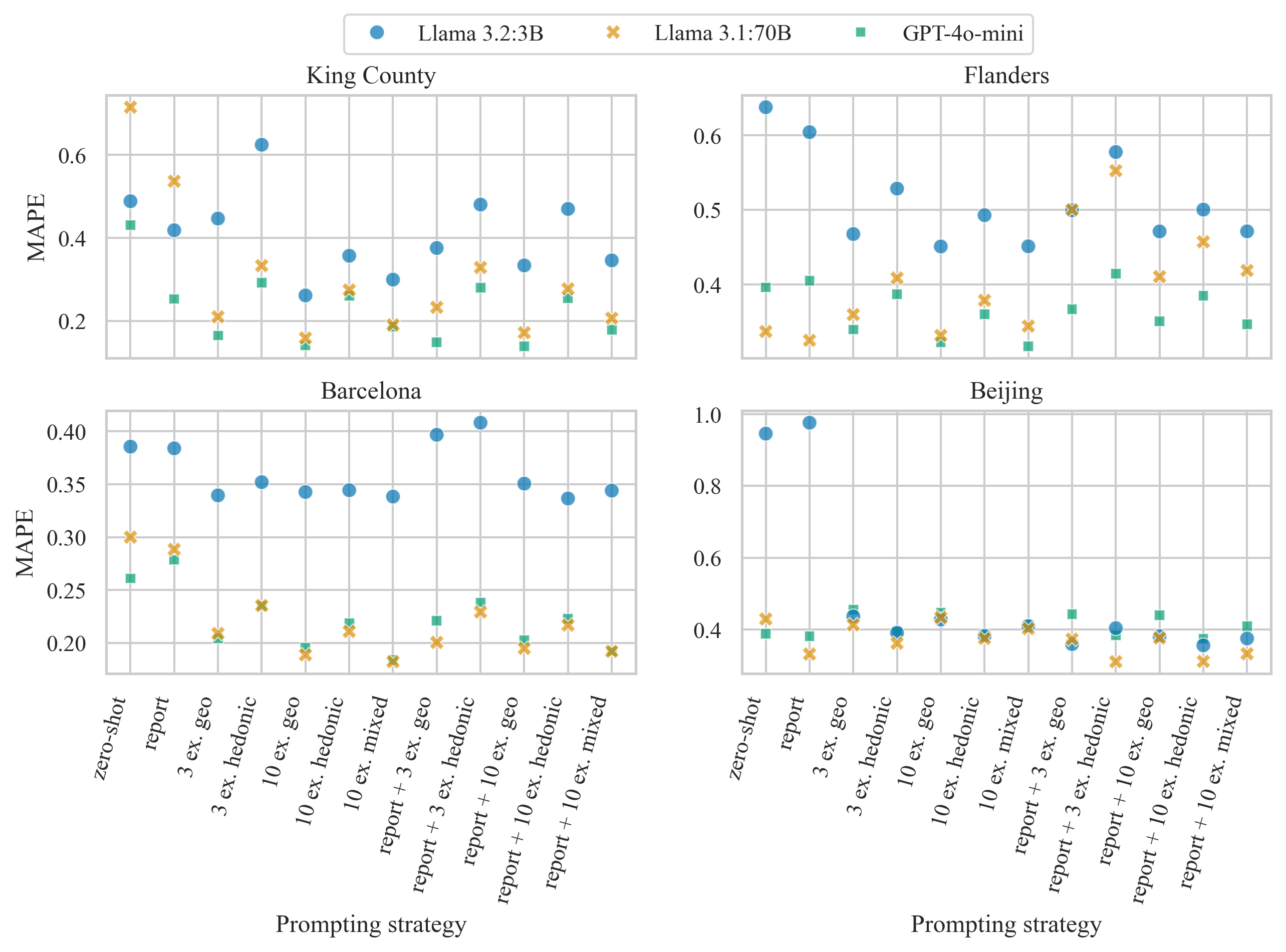}
    \caption{Including 10 mixed examples (geographic and hedonic similarity) provides the best results overall and zero-shot prompting the worst. GPT-4o-mini generally outperforms the other models. This figure shows the results for the twelve different prompting strategies across all four datasets.}
    \label{fig:mape}
    
\end{figure}

Figure \ref{fig:mape} summarizes the MAPE scores across all prompting strategies, models, and datasets. Generally, prompting strategies with more labeled examples lead to more accurate predictions, with ten mixed examples (\texttt{10 ex. mixed}) emerging as the best-performing strategy in most datasets and models. This suggests that combining geographically near and hedonic similar properties provides a balanced context that enhances LLM predictions. This approach aligns with prior work in hedonic price modeling, where integrating geographic and characteristic-based similarity has been shown to effectively capture local housing market dynamics \cite{Ozhegov2022}. Ten-shot prompting strategies consistently outperform three-shot prompts, independent of the selection method, indicating that LLMs benefit from comprehensive contextual information. Only for the Beijing dataset, the performance is sensitive to the specific method for example selection.

While \texttt{10 ex. mixed} performs best on average, the optimal strategy can vary by dataset. King County benefits from combining market reports with ten geographic neighbors (\texttt{report + 10 ex. geo}), Flanders and Barcelona favor ten mixed examples (\texttt{10 ex. mixed}), and Beijing performs best with market reports and three hedonic examples (\texttt{report + 3 ex. hedonic}). While a mixed example selection approach generally ranks best, geographic neighbors provide typically more useful contextual information than hedonic examples. Again, the Beijing dataset deviates from this observation and ranks hedonic examples higher. In most regions, spatial correlations in house prices are prominent, while Beijing's market may be influenced by higher levels of heterogeneity leading to property characteristics being more important in combination with broader economic trends. Another reason could be that the spatial structure in Beijing's house prices may be more complex than geographic proximity.

While market reports consistently improve zero-shot prompting, they are rarely effective on their own. Notably, in Beijing, monthly, city-specific reports ($\approx$ 501 characters in length) improve results across all prompting strategies, reinforcing the importance of market trends in this region and underscoring the value of fine temporal and spatial resolution. In the King County dataset, its US-wide, monthly market reports ($\approx$ 987 characters), which include quarterly sub-regional breakdowns of market trends, also benefit price prediction in most cases. In contrast, Barcelona and Flanders see limited or no gains: for both regions, quarterly, country-level reports ($\approx$ 1 817 characters for Barcelona, $\approx$ 1 566 characters for Flanders) are available, with Flanders additionally comparing neighboring countries and Barcelona highlighting only regions with the largest changes. This pattern suggests that higher temporal granularity (monthly vs. quarterly), tighter geographic specificity (city vs. national), and even concise report length materially enhance an LLM’s ability to incorporate dynamic market trends. Given that the King County and especially the Beijing dataset present significant temporal trends in house prices, these results show that market reports provide essential context for LLMs to capture temporal dynamics and thereby improve predictive performance.

Ranking the LLMs across datasets and prompting strategies reveals that GPT-4o-mini outperforms the other models, with Llama 3.1:70B as a close second. While GPT-4o-mini still achieves reasonable results with zero-shot prompts, Llama 3.2:3B consistently performs worst. This might be due to low-parameter models containing less world knowledge compared to high-parameter models, making them more dependent on prompts with richer context for accurate predictions. While larger models, Llama 3.1:70B and GPT-4o-mini, generally outperform smaller models, LLM performance also slightly varies across datasets. Llama 3.2:3B performs comparably to larger models on the Beijing dataset. This anomaly could be again due to the volatile market dynamics which makes all LLMs struggle equally. GPT-4o-mini performs best in King County and Flanders, whereas Llama 3.1:70B outperforms it in Barcelona and Beijing. This performance disparity could be attributed to GPT-4o-mini's stronger tailoring to the US and Western European contexts, where its training data and fine-tuning are more focused \cite{guo2024heygptracistanalysis}. In contrast, Llama 3.1, designed for multilingual text generation, appears to perform better for non-English or geographically diverse contexts, making it more adept at handling the varied linguistic and cultural features found in regions such as Beijing and Barcelona \cite{Meta2024}. Furthermore, GPT-4o-mini exhibits the least variability across prompting strategies, making it the most stable performer overall. While its advantage over Llama models is not always substantial, it consistently follows structured formatting, reducing output inconsistencies. Llama models, particularly Llama 3.2:3B, sometimes struggle to conform to the output format, which can contribute to higher prediction errors.

Despite some dataset-specific variations, \texttt{10 ex. mixed} emerges as the most robust prompting strategy, ranking the highest when averaged over models and datasets. This approach effectively balances geographic and hedonic information, making it a strong default choice when selecting prompting strategies for house price estimation. While market reports may further enhance performance in markets with strong temporal trends, their usefulness varies by region. Therefore, a hybrid approach—prioritizing mixed examples and potentially incorporating market reports when relevant—offers the most generalizable strategy.

\subsection{Positioning LLM Performance Relative to ML Baselines}
\subsubsection{Prediction accuracy}
Table \ref{tab:mape} compares the performance of LLMs with baseline methods using MAPE across datasets. The table shows the LLM results for the best-ranked prompt strategy \texttt{10 ex. mixed} and corresponding setting for kNN, alongside a SOTA GBT model with (LGBM) and without coordinates (LGBM $\varnothing$ XY). Generally, high-parameter LLMs outperform kNN, indicating they leverage labeled examples through ICL more effectively than kNN's simple interpolation. The Beijing dataset presents a particular challenge, with all LLMs performing worse than kNN. This is due to the strong temporal trends as established earlier, and can be mitigated with a market report (\texttt{report + 10 ex. mixed}) which results in a decrease in MAPE from 0.4022 to 0.3322 for Llama 3.1:70B, effectively outperforming kNN (0.3810).
\begingroup
\renewcommand{\arraystretch}{1.2} %
\begin{table}[ht]

    \centering
    \caption{LLMs generally outperform kNN and get competitive to SOTA models. Comparison of MAPE and PE Standard Deviation between LLMs with \texttt{10 ex. mixed} prompt and baseline models.}
    \begin{tabular}{c|cccc}
    \toprule
         & King County & Flanders & Barcelona&Beijing \\ \midrule

        Llama~3.2:3B & 0.2995 ± 0.3282 &0.4511 ± 0.5828 & 0.3383 ± 0.4211 & 0.4092 ± 0.1320\\
        Llama~3.1:70B & 0.1905 ± 0.2072 &0.3440 ± 0.4997 & \underline{0.1825} ± 0.2044	& \underline{0.4022} ± 0.1108\\
        GPT-4o-mini& \underline{0.1861} ± 0.1925 & \underline{0.3170} ± 0.4782 & 0.1842 ± 0.2004& 0.4125	± 0.1117\\ 
       
        \hline
        kNN & 0.2105 ± 0.2113  & 0.3207 ± 0.4380& 0.2638 ± 0.4070& 0.3810 ± 0.1292\\ 
        LGBM $\varnothing$ XY&  0.2391 ± 0.3220 & 0.3136 ± 0.4988 & 0.1936 ± 0.2825& 0.2427 ± 0.2031\\
        LGBM &\textbf{0.1378} ± 0.1611& \textbf{0.2625} ± 0.4170& \textbf{0.1556} ± 0.1780& \textbf{0.1056} ± 0.0840\\ \bottomrule

    \end{tabular}
    
    \label{tab:mape}
\end{table}
\endgroup

Comparing LLMs with LGBM $\varnothing$ XY, we see that LLMs show comparable performance. This indicates that LLMs are effective in extracting hedonic patterns from real estate pricing data. Finally, comparing LLMs with the SOTA LGBM models, we do not expect LLMs to perform better, but we see that LLMs can get relatively close without having access to the full dataset. In Flanders and Barcelona, GPT-4o-mini's MAPE is around 20\% higher than LGBM, while in King County the difference is 35\%. However, with the optimal strategy (\texttt{report + 10 ex. geo}), the MAPE in King County improves to 0.1390, completely matching LGBM's performance.

Interestingly, the Beijing dataset deviates in the baseline performance opposed to other datasets as well. It sees a great decrease in MAPE between kNN and LGBM $\varnothing$ XY, likely due to LGBM's ability to take advantage of temporal features. In addition, adding the geographic coordinates as predictors to LGBM reduces the MAPE from 0.2427 to 0.1056 for Beijing. While LLMs struggle with incorporating spatial relationships through neighboring examples, LGBM succeeds in deciphering the spatial structure in the dataset and significantly improves predictions. This strengthens our findings that LLMs can learn hedonic pricing patterns, but require more advanced techniques when the dataset is characterized by unconventional spatial structures and strong temporal dynamics.

 The PE Standard Deviation shows LLMs have error variability comparable to baselines, though Llama 3.2 exhibits greater fluctuation. Overall, LLMs surpass kNN and show competitive performance compared to SOTA models, particularly in extracting hedonic patterns from real estate pricing data. %

\subsubsection{Price Intervals}
To address \textbf{RQ3}, we assess LLM prediction intervals using the \texttt{10. mixed} prompt. Table \ref{tab:pi} reports two metrics: coverage (percentage of true prices within intervals) and MPIW (Mean Prediction Interval Width).
\begingroup
\setlength{\tabcolsep}{4pt}
\begin{table}[ht]
\caption{Prediction interval quality measured by Coverage (Cov.), percentage of true prices in test sample within intervals, and MPIW, Mean Prediction Interval Width. As we enforce LLMs to produce intervals around their predicted price, we included respectively 949, 872, 960, and 945 intervals for Llama 3.2:3B and 998 for Llama 3.1:70B on the Flanders dataset and 999 on the Beijing dataset.}
    \label{tab:pi}
    \centering
    \begin{tabular}{c|cc|cc|cc|cc}
    \toprule
        & \multicolumn{2}{c|}{King County} & \multicolumn{2}{c|}{Flanders} & \multicolumn{2}{c|}{Barcelona}& \multicolumn{2}{c}{Beijing}\\
         & Cov. & MPIW & Cov. & MPIW & Cov. & MPIW & Cov. & MPIW\\ \midrule
         Llama~3.2:3B &39.6 &220 289 & 36.7& 193 447& 46.8& 262 199 &\underline{10.8} & 1 625 475\\
        Llama~3.1:70B & \underline{57.5}& \underline{182 823}& \underline{51.4}&\underline{156 658}& \underline{64.0}&\underline{151 641}& 3.6&\underline{1 093 476} \\
        GPT-4o-mini& 35.5& \textbf{98 319}& 25.8&\textbf{65 488} &40.3& \textbf{74 444}& 1.2 & \textbf{514 394} \\ \hline
        LGBM & \textbf{90.5}& 316 293 & \textbf{90.5}& 317 476 & \textbf{86.2} &210 681 & \textbf{85.1}& 1 900 473  \\ \bottomrule

    \end{tabular}
    
\end{table}
\endgroup

LLMs generate narrower intervals but often miss the 90\% coverage target, showing overconfidence \cite{Xiong2023}. In contrast, conformal prediction enables LGBM to achieve near-target coverage but with wider intervals, illustrating the trade-off between coverage and precision. GPT-4o-mini produces the narrowest intervals but consistently underperforms on coverage, while Llama 3.1 offers the best balance across datasets. The Beijing dataset proves particularly difficult, with LLMs showing extremely low coverage and LGBM struggling despite conformal adjustments, likely due to the dataset's temporal trends. Despite adjusting for this distribution shift, this still influences predictions. LLMs may also lack geographical knowledge or show regional biases in Beijing \cite{manvi2024largelanguagemodelsgeographically}. Overall, it is clear that LLMs struggle with producing calibrated prediction intervals, but advanced techniques like conformal prediction or iterative prompting \cite{AbbasiYadkori2024,ye2024} that would be necessary to mitigate this problem, make it less evident for real estate practitioners to leverage LLM-based solutions.

\subsubsection{Feature Importance}
\begin{figure}[ht]
\centering
\includegraphics[width=.8\textwidth]{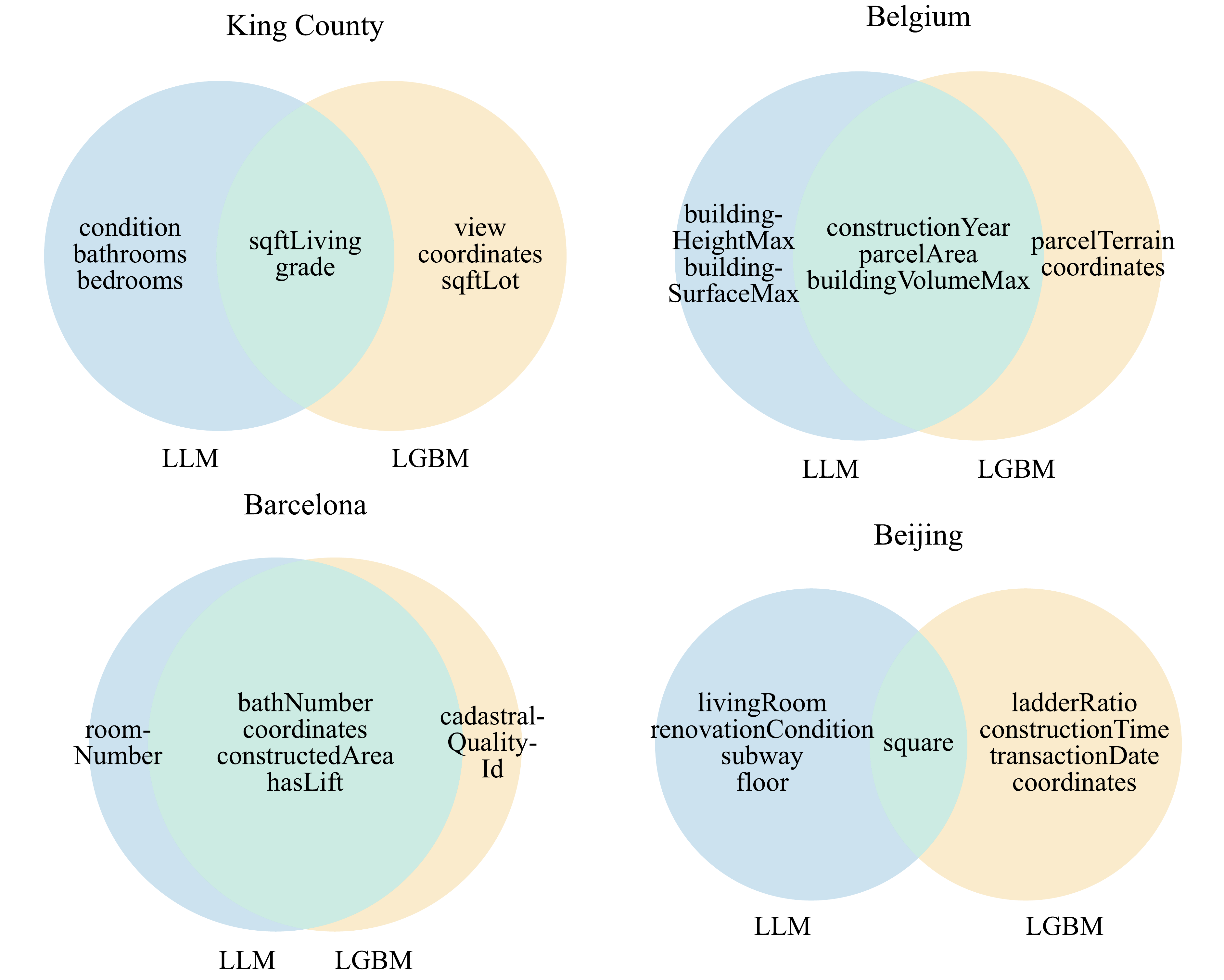}
\caption{LLMs generally align with LGBM on the importance of hedonic variables. Comparison of top five features between GPT-4o-mini and LGBM.}
\label{fig:features}
\end{figure}

We compare LLM-generated feature explanations to SHAP values from LGBM in Figure \ref{fig:features}, which shows the Venn diagrams of the top five features for all datasets. GPT-4o-mini generally aligns with LGBM on hedonic features, supporting their ability to extract property-related pricing patterns. This alignment suggests that LLM-generated explanations are not only consistent with established ML models but also offer a degree of trustworthiness, as they reflect key predictive drivers identified through robust, model-agnostic interpretability methods like SHAP. 

However, LGBM consistently ranks locational features, particularly coordinates, among its top predictors, while LLMs do not prioritize them, despite receiving full addresses and coordinates. The only exception is in the Barcelona dataset. This suggests that LLMs struggle with spatial reasoning, likely due to tokenization issues with coordinates and difficulty mapping addresses to house price patterns \cite{Li2024_spatialreasoning,manvi2024geollm}. Additionally, in accordance to the previous analyses, LGBM prioritizes temporal features in the Beijing dataset, while GPT-4o-mini does not. Similar to LLMs' issues with interpreting coordinates, LLMs might struggle with dates and 
 generalizing temporal trends in the data. Recent research has focused on improving temporal generalization of LLMs \cite{jin2023}. Although these limitations in spatial and temporal reasoning may explain the performance gap, caution is warranted since LLM self-explanations are not always reliable \cite{madsen-etal-2024-self}. Appendix \ref{a3} confirms the findings for Llama3.1:70b.

\subsection{Practical implications}
Our findings indicate that LLMs offer a low-barrier alternative to traditional ML solutions for real estate appraisal. With as few as ten examples, LLMs can provide reasonable price estimates, making them valuable for quick, accessible, and interactive valuations. We recommend structuring prompts with ten properties that are geographically near and share similar hedonic characteristics, while incorporating a market report in cases with strong temporal trends. 
LLMs are particularly suited for decision-support systems, especially for non-technical users and low-data environments. Private buyers and sellers, for instance, can use LLMs to gain insight into fair market prices, while creditors, real estate agents, and investors may rely on them for preliminary valuations that can later be refined with expert assessments or ML-based approaches.
\begin{table}[ht]
    \centering
    \caption{Comparison of LLMs and ML models for real estate appraisal}
    \label{tab:llm_vs_ml}
    \begin{tabular}{lcc}
        \toprule
        \textbf{Aspect} & \textbf{LLMs} & \textbf{ML Models} \\
        \midrule
        \multicolumn{3}{l}{\textbf{Accessibility}} \\
        \midrule
        Ease of Use & Works out of the box & Requires training \& tuning \\
        User Input & Natural Language via interface & Structured data through code \\
        \midrule
        \multicolumn{3}{l}{\textbf{Data Requirements}} \\
        \midrule
        Data Needs & Few examples needed & Large structured dataset \\
        Feature Handling & Implicit understanding & Manual selection required \\
        Unstructured Data & Can use reports \& text & Limited to structured inputs \\
        \midrule
        \multicolumn{3}{l}{\textbf{Model Capabilities}} \\
        \midrule
        Accuracy & Competitive, slightly below ML & State-of-the-art \\
        Geospatial Data & Limited handling & Explicitly modeled \\
        Temporal Trends & Struggles with time patterns & Can model time effects \\
        Prediction Intervals & Often overconfident & More calibrated \\
        Explainability & Text-based, intuitive & SHAP \& feature importance \\
        \midrule
        \multicolumn{3}{l}{\textbf{Practical Deployment}} \\
        \midrule
        Interactivity & Can take user feedback & Static predictions \\
        Computation & Free/low-cost web, API, local & Local CPU/GPU \\
        Best Use Case & Quick, flexible valuations & Accurate, large-scale modeling \\
        \bottomrule
    \end{tabular}
\end{table}

Table \ref{tab:llm_vs_ml} summarizes the key advantages and limitations of LLMs compared to ML models. While LLMs work out of the box, we identify three main limitations: they struggle with spatial reasoning, failing to properly integrate location effects; they exhibit weak temporal understanding, making it difficult to learn price trends over time; and they are overconfident in uncertainty estimation, often producing narrow and unreliable price intervals. Though market reports improve temporal generalization capabilities, fully addressing these weaknesses requires more advanced techniques, such as retrieval-augmented generation or fine-tuning \cite{Zhang2024BB}. These approaches introduce additional complexity that can undermine the accessibility and immediacy that make LLMs attractive alternatives to traditional ML methods. Similarly, reliable uncertainty estimation requires post-processing techniques that are challenging, especially with closed-source models \cite{AbbasiYadkori2024,ye2024}. Given these trade-offs, LLMs are best suited for fast, accessible valuations rather than high-accuracy, large-scale appraisals where calibrated uncertainty estimates are essential.

\section{Conclusion}
This study investigated how prompt engineering techniques optimize LLM performance for real estate appraisal (\textbf{RQ1}) and whether LLMs can serve as viable alternatives to traditional ML models (\textbf{RQ2}-\textbf{4}). Our results show that LLMs, when prompted using In-Context Learning with just ten real estate examples selected based on geographic and hedonic similarity (\textbf{RQ1}), can generate competitive price estimates (\textbf{RQ2}), making them a practical tool for real estate valuation. However, their spatial reasoning and temporal generalization capabilities remain limited, affecting the reliability of predicted price intervals (\textbf{RQ3}) compared to structured ML models. Nevertheless, LLMs align with ML models in explaining predictions based on property characteristics (\textbf{RQ4}), reinforcing their ability to capture hedonic valuation patterns.

By improving accessibility to property appraisal, LLMs help reduce information asymmetry in real estate transactions. While ML models remain more accurate in structured, large-scale applications, LLMs provide an interactive and intuitive alternative, particularly for non-technical users who need quick and interpretable price estimates. These findings align with \textbf{RQ2} and \textbf{RQ4}, highlighting that while LLMs can extract meaningful hedonic features, they require further refinement to fully capture spatial and temporal trends.

In summary, LLMs show potential for accurate and explainable price predictions. Future work should explore more recent LLMs with enhanced reasoning capabilities, alongside diverse prompting techniques such as chain-of-thought and self-consistency, or retrieval-augmented generation to further improve performance and robustness. Additionally, a systematic investigation into scaling LLMs with large-scale datasets is needed to provide deeper insights into data efficiency and generalization, though our results indicate that larger model sizes generally yield better accuracy. Exploring alternative geographic encoding strategies could also address current spatial reasoning limitations. Collectively, these directions offer a clear roadmap to enhance LLM trustworthiness and reliability, advancing their practical application in real estate appraisal.

\begin{credits}

\subsubsection{\discintname}
The authors have no competing interests to declare that are
relevant to the content of this article.
\end{credits}

\appendix
\section{Prompt template}\label{a1}
Table \ref{tab:template} provides the prompt template for house price prediction with LLMs including our prompt optimization strategies. 

\begin{longtable}[h]{p{4em}|p{8em}|p{21em}}
    \caption{Prompt template for house price estimation using an LLM. The structured prompt follows a multi-step approach to optimize predictions (step 1) and assess interpretability through uncertainty estimation (step 2) and feature importance (step 3).}
    \label{tab:template}\\
    
    Actor & Prompt Element & Content \\ \hline
     \endfirsthead
     Actor & Prompt Element & Content \\ \hline
     \endhead
        System &  System prompt & You are a real estate expert. \\ %
        \midrule
        \multicolumn{3}{ c }{Step 1: Price Prediction}\\
        \midrule
         \multirow{4}{*}{User} 
        &Task Definition &  Your task is to value properties in [region, country]. Properties will be described by a number of variables. Please use this information to predict the price in [currency]. Please answer with the price only and use the format `price [currency]'.\\ 
        &Market report (optional)&  The following report provides context on the housing market at the time of the transaction: [House Price Index report] \\
        &Examples (optional)& Here are some examples of relevant properties and their prices: The first property is located at [description of location and variables]. The transaction date is [transaction\_date] and the transaction price is [price] [currency]. The second property is located at [description of location and variables] ...\\
        &Property Details & Estimate the price, following the format, of a property located at [description of location and variables]. The transaction date is [transaction\_date]. The training data includes houses with prices ranging from [min\_price] [currency] to [max\_price] [currency], with a median price of [median\_price] [currency]. \\
        
        LLM & Price prediction & XXX [currency]\\ %
        \addlinespace
        \midrule
        \multicolumn{3}{ c }{Step 2: Uncertainty Estimation}\\
        \midrule
        User & Interval prompt & Please provide a price interval with 90\% coverage around your estimated price for this property in the format `min\_price - max\_price'. \\
        
        LLM & Price interval & XXX - XXX \\ %
        \midrule
        \multicolumn{3}{ c }{Step 3: Feature Importance}\\
        \midrule
        User & Feature prompt & Please provide the top 5 features that you deemed most important for your previous predictions. Answer with a comma-separated list of features, using the feature names: [variables].\\
        LLM & Feature ranking & $feat_1, feat_2, feat_3, feat_4, feat_5$ \\

\end{longtable}

\section{LLM implementation details}\label{a2}
More information on the checkpoints and computing environment used for the experiments with LLMs can be found in Table \ref{tab:llm}.
\begin{table}[h!]
    \centering
    \caption{Implementation details of LLMs}
    \label{tab:llm}
    \begin{tabular}{c|c|c}
    \toprule
        Checkpoint & Access point & Compute\\ \midrule 
        \texttt{Llama3.2:3b-instruct-fp16} & Ollama\footnotemark{} & Google Colab T4 GPU \\
        \texttt{Llama3.1:70B-instruct-fp8}\footnotemark{} & vLLM \cite{kwon2023efficient} & NVIDIA A100 \& H100 GPUs\\
        \texttt{GPT-4o-mini-2024-07-18} & OpenAI API & - \\  
        \bottomrule
    \end{tabular}
\end{table}
\addtocounter{footnote}{-2}
\stepcounter{footnote}\footnotetext{\url{https://ollama.com/}}
\stepcounter{footnote}\footnotetext{quantized by Neural Magic (\url{https://neuralmagic.com/})}

\section{Feature importance}\label{a3}
\begin{figure}
    \centering
    \includegraphics[width=0.9\linewidth]{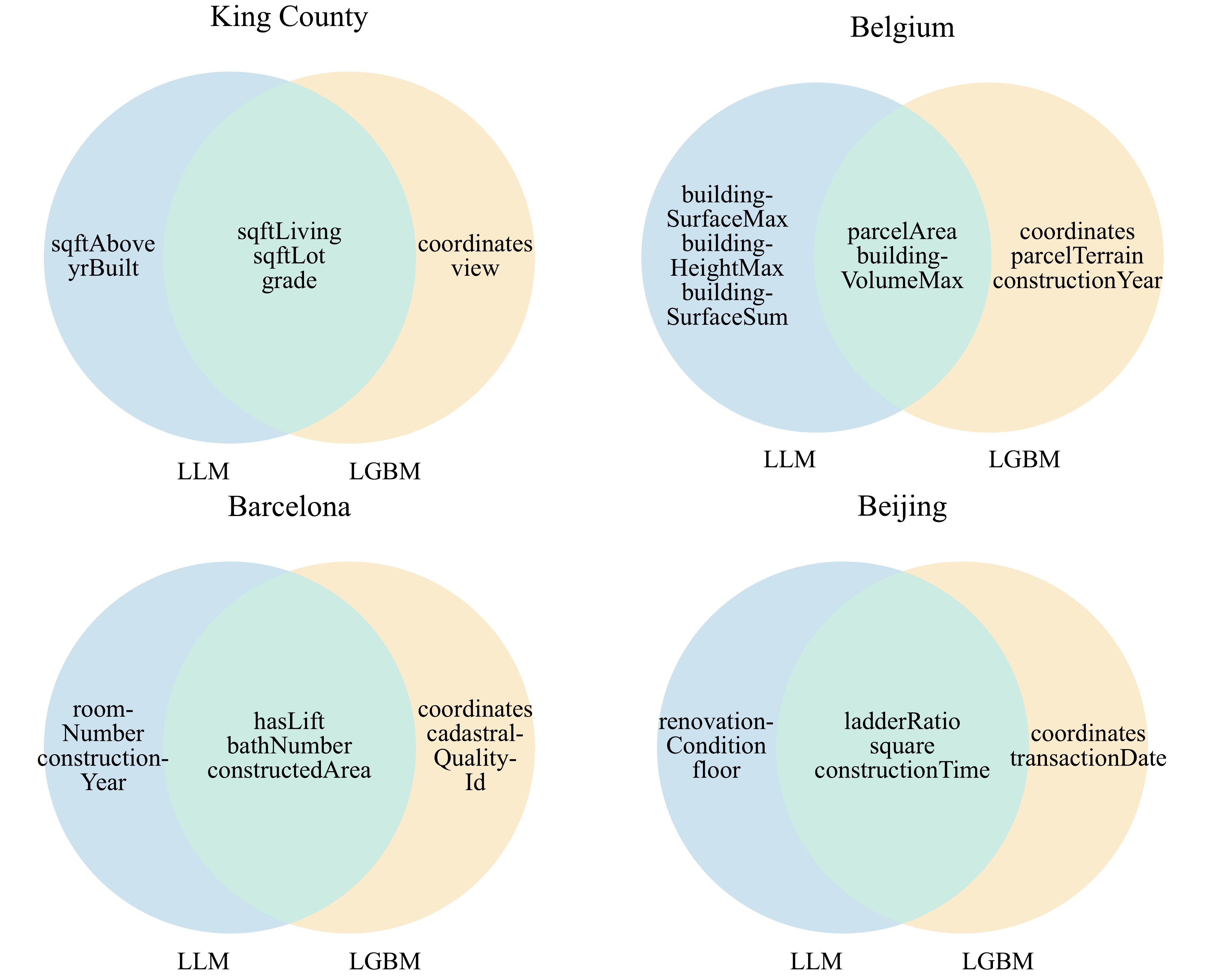}
    \caption{LLMs generally align with LGBM on the importance of hedonic variables. Comparison of top five features between Llama3.1:70b and LGBM.}
    \label{fig:features2}
\end{figure}

Venn diagram of the feature rankings for Llama3.1:70b versus LGBM is depicted in Figure \ref{fig:features2}. The findings align with those based on the results from GPT-4o-mini: LLMs align with LGBM on hedonic features, while coordinates and transaction date, representing spatial and temporal dynamics, are only picked up by LGBM.

\end{document}